\documentclass[sigconf]{acmart}

\AtBeginDocument{%
  }

\usepackage{lipsum}
\usepackage{algorithm}
\usepackage{algcompatible}
\usepackage{algpseudocode}
\usepackage{array}
\usepackage{textcomp}
\copyrightyear{2024} 
\acmYear{2024} 
\setcopyright{rightsretained} 
\acmConference[ICAIF '24]{5th ACM International Conference on AI in Finance}{November 14--17, 2024}{Brooklyn, NY, USA}
\acmBooktitle{5th ACM International Conference on AI in Finance (ICAIF '24), November 14--17, 2024, Brooklyn, NY, USA}
\acmDOI{10.1145/3677052.3698653}
\acmISBN{979-8-4007-1081-0/24/11}

\begin{document}

\title{WallStreetFeds: Client-Specific Tokens as Investment Vehicles in Federated Learning}

\author{Arno Geimer}
\authornote{Corresponding author.}
\email{arno.geimer@uni.lu}
\orcid{0000-0002-4442-4847}
\affiliation{
  \institution{SnT, University of Luxembourg}
  \city{Luxembourg}
  \country{Luxembourg}
}

\author{Beltran Fiz}
\email{beltran.fiz@uni.lu}
\orcid{0009-0006-7171-3253}
\affiliation{
  \institution{SnT, University of Luxembourg}
  \city{Luxembourg}
  \country{Luxembourg}
}

\author{Radu State}
\email{radu.state@uni.lu}
\orcid{0000-0002-4751-9577}
\affiliation{
  \institution{SnT, University of Luxembourg}
  \city{Luxembourg}
  \country{Luxembourg}
}

\begin{abstract}

Federated Learning (FL) is a collaborative machine learning paradigm which allows participants to collectively train a model while training data remains private. This paradigm is especially beneficial for sectors like finance, where data privacy, security and model performance are paramount. FL has been extensively studied in the years following its introduction, leading to, among others, better performing collaboration techniques, ways to defend against other clients trying to attack the model, and contribution assessment methods. 

An important element in for-profit Federated Learning is the development of incentive methods to determine the allocation and distribution of rewards for participants. While numerous methods for allocation have been proposed and thoroughly explored, distribution frameworks remain relatively understudied.

In this paper, we propose a novel framework which introduces client-specific tokens as investment vehicles within the FL ecosystem. Our framework aims to address the limitations of existing incentive schemes by leveraging a decentralized finance (DeFi) platform and automated market makers (AMMs) to create a more flexible and scalable reward distribution system for participants, and a mechanism for third parties to invest in the federation learning process.

\end{abstract}


\maketitle

\section{Introduction}\label{sec:intro}

Data has emerged as a critical asset in today's digital economy. It drives innovation and enables the rapid advancement of machine learning algorithms. As organizations increasingly rely on data to gain competitive advantages, the ability to collect, analyze, and create collaborative environments has become a significant challenge. Federated Learning (FL) addresses this issue by allowing multiple participants to collaboratively train a machine learning model without the need to share their raw data.Originally conceived by \cite{mcmahan2017communication} at Google Research, in a classic FL process involves multiple rounds of learning, where participants train a local model using their own data and then share model updates with a central server, where they are aggregated.  One of the  key advantages of FL is that by keeping data on the devices, sensitive information remains localised, mitigating the risks associated with the new attack vectors a centralised solution could expose all participants to, such as data breaches and unauthorised access. Additionally, FL models often exhibit superior generalizability compared to their centralised counterparts \cite{huang2023federatedlearninggeneralizationrobustness}.

The financial sector, with its stringent data privacy regulations and sensitive information, stands to benefit immensely from FL. Applications such as such as fraud detection \cite{ijcai2020p642}, credit risk assessment \cite{lee2023federated}, and compliance with regulatory frameworks \cite{ORBi-edc319d4-e718-4c9c-9747-048ec8f6bae2}  can be significantly enhanced through secure and collaborative model training. However, the adoption and sustained success of Federated Learning in finance will depend on overcoming unique challenges related to privacy, security, and participation incentives.

Privacy and security have  been a core focus in FL research from its inception. Early work focused primarily on deploying differential privacy techniques to protect individual data contributions within the aggregated model updates  \cite{DBLP:journals/corr/abs-1911-00222}, later evolving to encompass a larger scope and attempting to safeguard the entire training process, such as secure multi-party computation and homomorphic encryption \cite{bonawitz2016practicalsecureaggregationfederated}. 

From the economic perspective, the incentives for being in a federation are crucial to motivate client participation. While traditionally the benefits of the global model are shared amongst the participants, the improvements from the global model might not be enough to justify the capital investment required to partake in the federation. In fact recent studies have shown that one of the primary challenges to lasting federated learning partnerships is the initial upfront costs and incentives to remain in a federation for long durations \cite{bi2023understanding}. 

Blockchain technology and decentralised finance (DeFi) have revolutionised revenue generation models for new companies. By leveraging decentralised ledgers, new businesses can create and manage digital tokens while smart contracts automate and secure transactions, reducing operational costs. These  can be offered in the form of Initial Coin Offerings (ICOs), providing early-stage funding without traditional venture capital constraints. The transparency and security inherent in blockchain technologies also serve to attract a wider range of investors. 

Our research aims to addresses the gap in participation incentives by leveraging on the infrastructure in place provided by DeFi platforms. We propose a federated learning framework that allows participating clients to emit a series of tokens to represent their contributions to a federation's learning process. In doing so, we are able to reduce the barrier to entry for data producers by providing them with alternate means to raise capital. 

In this work, we introduce a novel reward payout scheme that enables federated learning participants to monetize their future contributions to a collaboratively trained model.
Additionally, we propose an innovative investment vehicle for external actors that emphasizes the learning effectiveness of participants' data.

This paper is structured as follows: Section \ref{sec:background} provides a background on Federated Learning and DeFi. Section \ref{sec:related} goes over related work relevant to our proposed framework. In Section \ref{sec:restrictions}, we discuss the limitations of current incentive schemes. Section \ref{sec:tok} introduces our tokenized reward-sharing framework, followed by an evaluation and discussion. Finally, Section \ref{sec:con} concludes the paper with some final thoughts and future research directions.

\section{Background}\label{sec:background}
This section provides an explanation of Federated Learning as well as its application in different areas. We take a closer look at payment types and various reward distribution and contribution techniques.

\subsection{Federated Learning}
Federated Learning is a collective machine learning technique allowing privacy-preserving training of a joint model. A federation consists of a central authority, or server, and a collection of clients. After deciding on a model type and parameters, the federation framework and the common task, the central server initiates a model and distributes its weights to the participants. These train the model on their respective dataset and set the updated weights back to the server. The server aggregates the weights, based on a previously decided aggregation strategy, and returns the new model to the clients. This process is iterated until convergence is reached. 

Algorithm \ref{alg:FL} showcases a standard Federated Learning process. ClientUpdate represents the specific training process used by the clients, which involves minimizing the loss function.
ServerUpdate is the aggregation function employed by the central server. It is heavily influenced on the needs and type of federation. Aggregations designed to reduce communication costs by only polling model updates from a few clients have been proposed \cite{DBLP:journals/corr/abs-2010-13723}. Other techniques speed up convergence by giving more weight to specific clients, or by using superior optimization algorithms \cite{reddi2020adaptive}\cite{hsu2019measuring}. More stable convergence can be obtained by changing the aggregation function type \cite{yin2018byzantine}.

\begin{algorithm}[ht]
\caption{A Federated Learning framework with $K$ clients.}
    \begin{algorithmic}[1]
    \Statex{\textbf{On server:} Initialize $\omega_1$}
        \For{each round $t = 1, 2, ..., T-1$:}
            \State{\textbf{On server:}}
                \Statex{\hspace{0.75cm}Distribute $\omega_t$ to the clients}
            \State{\textbf{On client $k$}:}
            
                \Statex{\hspace{0.75cm}$\omega_{t+1}^{k}$ $\gets$ ClientUpdate($k$, $\omega_t$)}
                \Statex{\hspace{0.75cm}Send $\omega_{t+1}^{k}$ to the server.}
            \State{\textbf{On server:}}
            \Statex{\hspace{0.75cm}$\omega_{t+1} \gets $ ServerUpdate($\omega_{t+1}^1 , ... , \omega_{t+1}^K)$}
            
        \EndFor
    \State{The \textbf{final model $\omega_T$} is kept as the federated model.}
    \end{algorithmic}
\label{alg:FL}
\end{algorithm}
\subsubsection{Federation types}
While the FL process stays the same across the board, due to differences in setups and variables such as the number of clients and the amount of data, two different types of federations have emerged:

\textbf{Cross-device} federations contain a vast number of clients with sparse datasets. Computational capacities on participants are usually limited. Examples might include IoT devices \cite{nguyen2021federated} , satellites \cite{matthiesen2023federated} and electrical meters \cite{taik2020electrical}. The primary challenge is motivating the large number of participants to join and remain in the federation despite the limited computational resources and potential costs related to energy consumption and network usage \cite{kang2019incentive}. 

\textbf{Cross-silo} federations consist of few participants, typically companies or institutions with extensive datasets, such as hospitals, banks and and similar organizations. In these federations, communication costs and computational capabilities are of minor importance, since the participating entities are usually well equipped to handle these problems. However, as participants may be competing entities, privacy preservation as well as security from malicious attacks are of utmost importance. It has been shown \cite{geiping2020inverting} that in federated settings without adequate security measures, nosy participants can gain information about other clients' datasets and even singular data samples. Even though a federated model is a net benefit to most actors, privacy concerns may lead to companies refraining from joining a federation. Thus, besides privacy assurances, incentives play a critical role in convincing participants to join a cross-silo federation\cite{kairouz2021advances}. Although they can play a role in cross-device federations \cite{kang2019incentive}, incentive mechanisms are more commonly applied in cross-silo settings.

\subsubsection{Model types and application domains}
Federated Learning was introduced as a method of collaboratively training convolutional neural networks for image recognition and LSTMs for language model. Since then, Federated Learning has seen usage in a wide are of domains. Those include tabular data modeling using federated versions of XGBoost \cite{yang2019tradeoff}, used by WeBank among others \cite{liu2021fate}. Privacy issues have been adressed with the inclusion of homomorphic encryption into Federated Learning\cite{zhang2020batchcrypt}.
\subsection{Smart Contracts}

Blockchain based smart contracts became popular through their deployment in the Ethereum blockchain \cite{buterin2013ethereum}. They are executed by validator nodes with the assurance of the underlying consensus protocol, verifying and enforcing the terms of the smart contract directly in a trustless and transparent manner. The currency used is Ether, which can be utilized to transfer wealth when sent to other accounts or to execute a smart contract. These smart contracts are usually developed in a higher-level language like Solidity, which is then converted to bytecode and stored on the blockchain. This bytecode is interpreted by the Ethereum Virtual Machine (EVM), which executes the smart contract whenever a transaction interacts with the contract account.
 
 Smart contracts have paved the way for a wide range of decentralized applications that leverage the immutable and decentralized nature of the blockchain. For instance, in the financial sector, smart contracts are used to create DeFi platforms that offer services like lending, borrowing, and trading without traditional financial intermediaries.

\section{Related Work}\label{sec:related}
We now explore the work related to our proposed framework. We investigate incentive types in Federated Learning, how contributions may be calculated, and we introduce Automated Market Makers, Decentralized Finance-based exchange platforms.

\subsection{Federated Learning Incentive frameworks}

\subsubsection{Reward Types}
While usually not mentioned as a specific aspect of FL in literature, different types of reward have emerged depending on the type of setting and federation. Most common are value-based rewards, either monetary or cryptocurrency-based\cite{pandey2022fedtoken}, arising in federations where the central server owns the model and wants to use data owners to improve its performance. These rewards allow participants to be paid before, during and/or at the end of the federation run. Incentive mechanisms form the core dynamics of reward distribution between the server and clients.

\subsubsection{Client-pricing incentive mechanism}
Client-pricing mechanism allow clients to sell model updates to the server.
Since clients are independent entities who make their own decisions, game theory can be used to either maximise the overall payoff of the federation or that of individual clients. Depending on the nature of the federation, clients may use different strategies:

During a \textbf{coalitional game}, clients cooperate to maximise the global model performance, entering into contracts between each other and coordinating strategies. 

In a \textbf{Stackelberg game}, specific clients, the followers, wait for other clients, the leaders, to make their move before deciding on a pricing strategy. The leaders in a Stackelberg game choose their strategy to maximize their profit, while followers aim to gain the maximum utility given the leaders' chosen strategy.

Within a \textbf{Non-cooperative game}, players do not care about the outcome of the federation, they are only interested in maximizing their own profit.

An \textbf{auction-based} incentive scheme \cite{9372882} \cite{zeng2020fmore}comprises clients determining the cost of being involved in training, and making bids for their participation to the central server. The central server then determines winners, includes them in training, and pays them when finished. Auctions can take on different forms\cite{tu2022incentive}, including \textit{open-cry auctions} where bids are public and \textit{sealed-bid auctions} where bids are private.

A \textbf{performance-based} incentive mechanism was introduced in \cite{DBLP:journals/corr/abs-1906-01167}. The authors' protocol encourages participants to  exchange model updates for tokens, which they can use to buy other model updates for aggregation. A participant who regularly sells their model update up for trade will collect more tokens and attain a better performing final model.

\subsubsection{Server-based incentive mechanisms}

Opposed to these client-pricing incentive mechanisms are server-based incentive mechanisms, in which the central server distributes a value-based reward to participants. Here, clients do not price and sell their involvement in the federation, they get paid to be part of the federation. How the reward is distributed is to be decided during the setup of the federation: 

A \textbf{linear} reward model pays out based on a participant-specific quantity, such as the dataset size of participants. This value might already be known by the central server, as it is used in aggregation strategies as a means to weighting model updates when aggregating.

An \textbf{equal} reward model hands out equal rewards to each participant. While this settles disagreements about the fairness of contribution calculations between participants, it may discourage potential substantial clients from joining, as the federation might not be able to cover their costs.

A \textbf{contribution-calculation based} incentive model has the server calculate the contribution of a client to the federation and distributes rewards based on this assessment. A basic method of contribution assessment would be the performance of clients' model updates.
However, more extensive contribution methods have been proposed, of which the most popular are Shapley values \cite{shapley1953value}.
Given a federation with a set of clients $C$, the Shapley value $\phi_i$  of client $i$ can be calculated as \[\phi_i = \sum_{S \subseteq C \setminus {i}}{\frac{|S|!(C - |S| -1)!}{C!} (f(\omega_{S \cup i}) - f(\omega_{S}))},\] where $f$ is the utility of the model, usually the accuracy or loss on a test set, and $\omega_{S}$ is the combined model of the subset of clients $S$ \cite{song2019profit}. Shapley values do necessitate a local test set on the server, and additionally introduce a substantial computational effort, as the amount of per-value computations increases exponentially with the number of clients.

\begin{figure}[h]
    \includegraphics[scale = .6]{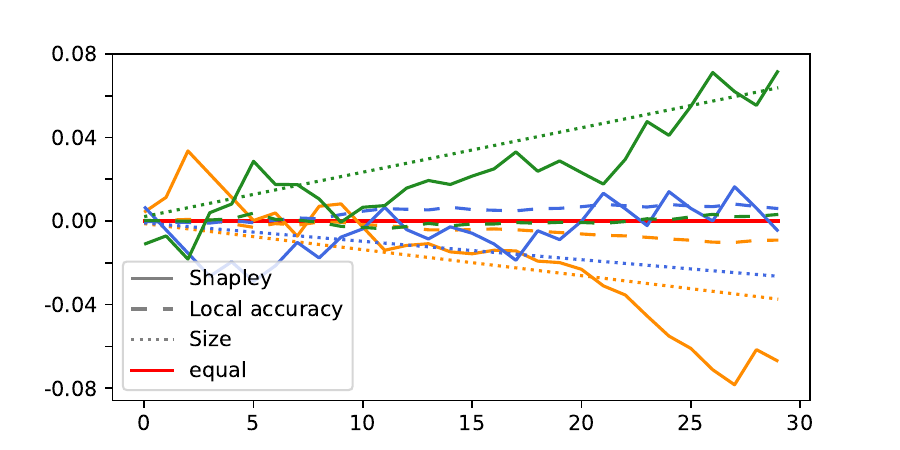}
    \caption{Per-client percentage of reward received as compared to equal payouts, with various contribution calculation methods, on CIFAR-10. Colors indicate different clients.}
    \label{img:icaif-prices}
\end{figure}

Figure \ref{img:icaif-prices} showcases the cumulative per-round percentages of rewards using different contribution calculations, compared to an equal payout which distributes the payment evenly between all clients each round. As can be seen, client rewards are more volatile when using Shapley values, allowing clients to collect more rewards depending on the quality of their model updates. 

However a contribution technique like Shapley values would accurately reflect utility differences in non-independent and identically (Non-IID) distributed data sources.

\subsection{Automated Market Makers}
On a traditional exchange market, buyers and sellers can place orders to trade assets. The order book is public information, allowing third parties to gain an insight into the valuation and demand of assets. To match orders, market makers manage the exchange, profiting off the spread. However, in markets with low liquidity, poor availability of assets leads to wide spreads, preventing market makers to properly match orders, resulting in no trade.
Thus, in Decentralized Finance (DeFi) applications, automated market makers (AMM) were proposed. Instead of requiring traders to set buy and sell orders and managing the spread, AMMs determine prices algorithmically, allowing for faster and smoother execution of trades.

The most basic type of AMM allows for the exchange of two currencies \cite{mohan2022automated}. 
Liquidity providers offer a supply of tokens to the AMM, which are pooled and used in any future trades. In such a two-currency AMM, let the currencies be $X$ and $Y$, with $x$ and $y$ the amount of each currency held by the market maker. The AMM's pricing algorithm determines the value of an asset based on the trading function $f(x, y) = k$ for an arbitrary predetermined value $k$. Some examples of such trading functions are the multiplicative trading function $f(x, y) = x * y$ or the additive function $f(x, y) = x+y$. When a trader wants to exchange $X$ for $Y$, the trading function determines how many tokens of $Y$ the trader receives. Meaning, if a trader wants to exchange $m$ amount of $X$ for $n$ amount of $Y$, the trade is accepted if $f(x, y) = f(x+m, y-n) = k$.
It follows that demand drives prices, as an increase in buyers leads to an increase in availability of a token, which results in a higher exchange rate.

Constant mean market makers (CMMM) allow for trading of multiple assets on one single exchange \cite{angeris2022constant}. Their trading function $f$ is the geometric mean of the amount of their $n$ held currencies, i.e. $f(x_1, ..., x_{n}) = \prod_{i=1}^{n}{x_{i}^{\omega_i}}$, with $\sum_{i}{\omega_i} = 1$. Currency $n$, the baseline currency, is called the numeraire. It is usually denoted in US Dollars ($\$$ or USD) or Euros (€ or EUR), and its value is always 1.
Initial values are reflected in the trading formula, and determined at the creation of the AMM.
Ultimately, the AMM scalability is limited by the underlying blockchain protocol \cite{xu2023sok}.

In summary, an automated market maker owns a pool of currencies, and facilitates trading by using a trading function to determine exchange rates. This effectively reduces friction and enhances the overall liquidity of the platform.

\section{Limitations of current incentive schemes}\label{sec:restrictions}

Although current Federated Learning incentive frameworks introduce state of the art valuation mechanisms to FL, the implementation and execution of payouts is understudied. Incentivizing participants to join and stay in the federation is still challenging due to, among others, the inflexibility of current payout mechanisms. This section goes into more details on which problems may arise and their consequences.

\subsection{Closed flow of capital}
\begin{figure}[h]
\includegraphics[width=6cm]{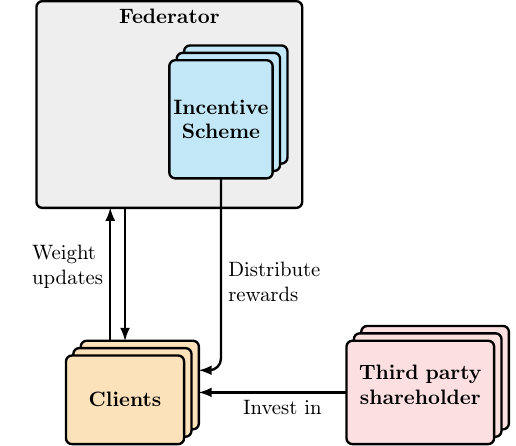}
\caption{An illustration of incentive schemes, closed systems where third party investments have to be be negotiated directly with the client.}
\label{fig:curr_inc}\end{figure}

All of the previously mentioned value-based incentive mechanisms are closed systems, as shown in Figure \ref{fig:curr_inc}: on the basis of a chosen incentive mechanism, monetary transactions are executed directly between the central server and clients. External backers need to enter into a funding agreement with a client or the server directly, which leads to regulatory challenges, necessitating agreement talks and setting up contracts. Such closed incentive schemes severely restrict the inflow of capital to the federation.

Lack of investment can have severe consequences, not only for the participants, but for the federation as a whole: Being underfunded, clients may drop out of the federation, leading to the loss of valuable information gained from their data. The shared model decreases in performance, and loses in value, which diminishes rewards. This can result in a chain reaction: Other clients are not satisfied with rewards anymore and drop out themselves, leading to a further performance loss. In the end, the federation either collapses, or only includes a few clients with a badly-performing shared model. On the other hand, outside investment allows clients to acquire more data and better computing resources in the hopes of receiving more rewards, which improves global model performance.

\subsection{Upfront cost}
A well-known problem in Federated Learning are the upfront costs encountered when joining the federation: Participants may not own the computational equipment required to carry out the calculations necessary in a Federated Learning environment. To deal with this problem, incentive schemes in which the server pays out a portion of the total rewards in the beginning have been proposed\cite{yu2020sustainable}. However, in such a case, the server needs to have the financial means to pay clients upfront, which may not be the case. Additionally, it is not sure whether a client will generate enough income during the federation to justify the initial investment. While a more flexible reward scheme is in the interest of the federation, we argue that opening up the flow of capital by allowing third parties to get involved takes the burden of financing clients off the server, and leads to a more efficient federation in general. 

\subsection{Consequences}

The closed structure of the federation disallows direct outside investment. A third party interested in the use case, who does not own any training data on their own, has no prospect of profiting off the federation without directly investing in clients. This means entering into direct negotiations with investors, a long and tedious process, in which both parties need to agree on the amount to invest, the specific usage of the money, as well as the return. Such a tedious process can discourage investors.

Due to strict privacy regulations, attracting clients to a federation is already a problem in Federated Learning \cite{bi2023understanding}, and potentially useful datasets stay unused. Additionally, current reward distribution systems do not allow for enough versatility to incentivize some types of participants.

As an example, tech startups amass vast amounts of private data. As a result of to their exceptionally high failure rate, many startups have to cease operations while sitting on a big pile of unused data. This data could still be monetized by participating in federations and collecting rewards. However, steep upfront costs, coupled with payment inflexibility, make it extremely difficult to join the federation. Due to data privacy issues, there is no possibility for the startup to sell out its data to the federation upfront. The data is not lost because the company does not want to participate; the data is lost because the company is not able to monetize its data upfront.

In summary, two types of problems emerge in Federated Learning incentive mechanisms: On the one hand, clients have difficulties in raising capital, either to enter the federation, or to acquire new data sources. On the other hand, outside investing in the federation is not encouraged, and is only possible by directly funding participants or the server. 

A solution would be for participants to go public: By issuing shares in the company, they raise funds, and third party investors can directly invest. However, there are several reasons for staying private: Companies keep more control over their operations, do not have to adhere to compulsory regulations and avoid the risks associated with an IPO. 

For these reasons, a mechanic to monetize participation and expected future rewards at any point in time is sensible. Enabling participants to trade expected reward payments to raise capital from the start, helps them fund operations and extend their business. On the other hand, this allows outside investors to exchange capital for the opportunity to profit off potential growth in rewards.

In the following sections, we will introduce and discuss a novel reward distribution platform for Federated Learning and demonstrate how it can mitigate the issues presented here.

\section{Tokenized reward sharing framework}\label{sec:tok}

As showcased in \ref{sec:restrictions}, a standard continuous Federated Learning incentive mechanism consists of a determining of the value of a participant's model update and a subsequent exchange of this update for a monetary reward between the client and the server at each round. The determination of the price is either left to the participant or the central server. This comes with several challenges, including fair pricing and the inflexibility of reward distribution leading to financing problems. 
We propose a tokenized reward distribution mechanism for Federated Learning, allowing participants to sell stakes in their involvement in the federation to outside actors.

\subsection{Architecture}

In a federated learning environment, the only flow of capital are the payouts of rewards from the central server. The revenue of the federation is contained in a closed system, which does not allow outside parties to profit directly.
That is what we propose as the basis of our framework: In exchange for  investing in a client's operations, third parties receive parts of the rewards received by said client. This is a very similar situation to public companies, where investors can receive dividends from a company's profits as a return on their investment.

\begin{figure}[h]
\includegraphics[width=8cm]{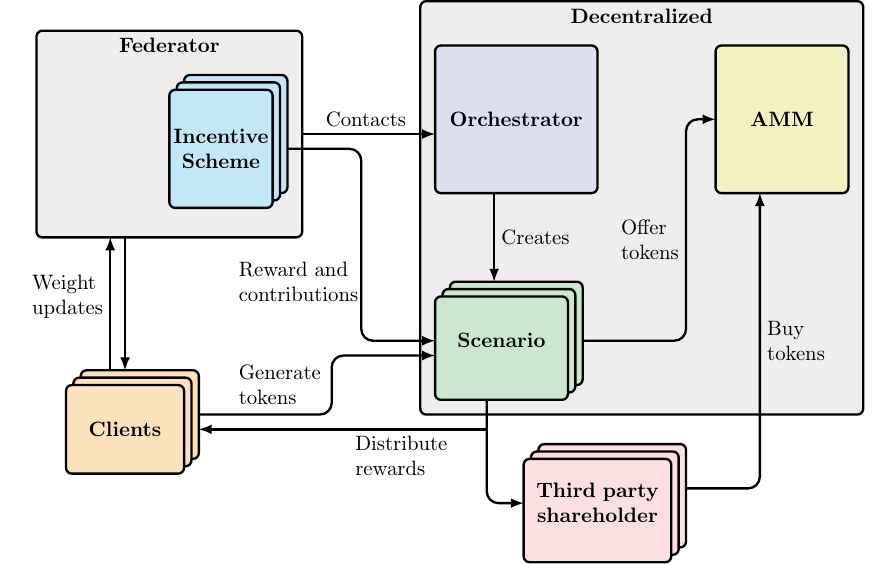}
\caption{Our share-based reward allocation platform allows payments to go from the server directly to shareholders. Here, the client still holds some shares, as they obtain a part of the reward.}
\label{fig:sha_exc}
\end{figure}

However, the implementation of such a framework in Federated Learning raises some difficulties. A third party investor would still need to invest money directly in a client, which ties them to all of the client's operations. Determining the return for an investor involves negotiations with the client. Additionally, investors can only make fair value assessments of different clients if reward payouts are public information. We use decentralized blockchain technologies, namely smart contracts and tokens, to open up the reward sharing framework: All privacy-sensitive operations are handled by the main server, but everyone can view the distribution of rewards.

More specifically, as shown in Figure \ref{fig:sha_exc}, our platform introduces three new parties as a decentralized addition to the federation: An \textbf{Orchestrator}, a \textbf{Scenario} and an \textbf{Automated market maker} (AMM). The Orchestrator is a smart contract, which can be contacted by a interested Federator to create a Scenario. The Scenario, a smart contract created by the Orchestrator for the specific task, manages the process of onboarding clients and acts as the central hub for reward distribution. While the Orchestrator and AMM can be created and owned by the Federator, they may be independent entities, and as such, would rely on themselves for funding. As decentralization is a desirable trait in Federated Learning we treat both, and by extension also the Scenario, as decentralized standalone entities.

Paying rewards to holders of tokens opens up the incentive scheme: If a third party wants to get involved in a client's operations in the federation, they can purchase tokens and collect a return on their investment via the rewards paid out to the token holders. However, if tokens are directly sold to investors, the system is still closed, as there is no open market for buying and selling. 

Thus, we introduce an automated market maker (AMM) to the decentralized part of the federation. As a marketplace for the selling and buying of tokens, the AMM facilitates the exchange and enabling speculation on token prices due to potential future rewards.
More specifically, we propose to use a constant mean market maker (CMMM), enabling the exchange of multiple tokens as well as trading for a base currency. This base currency allows any party, be it outside investors or participants, to purchase tokens, and thus involve themselves in a specific client's operations in the federation. The AMM ensures liquidity at any moment, facilitating trading. it is not tied to a specific federation: Indeed, an AMM allows for any token to be listed. Thus, the AMM makes it possible for third parties to invest in different federations, in different clients, as well as in the same client across different federations. Transactions are public: Anyone has insight into token prices and ownership. This transparency enables investors to estimate values, allowing them to determine potentially over- or undervalued clients, and to invest their money accordingly.

Our framework does not depend on a specific incentive scheme: A client-pricing incentive mechanism can just as easily be applied as a server-based incentive mechanism. As long as a flow of capital from the federator to individual clients takes place, the framework can be employed to enable third-party investment.

\subsection{User journey}
We describe the user journey, from the creation and joining of a federation to the payment of rewards, in more detail.

\begin{figure}[h]
\includegraphics[width=8cm]{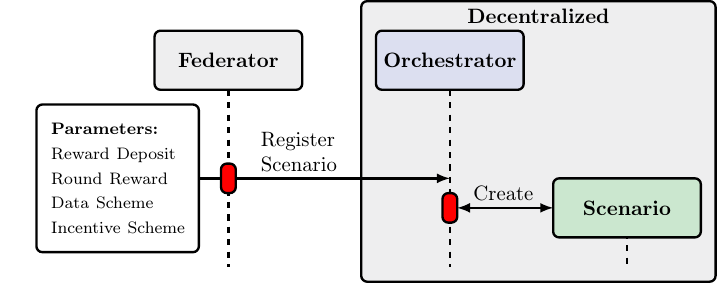}
\caption{The creation of a specific Scenario.}
\label{fig:j1}\end{figure}

The Federator, or central server, registers the federation on the Orchestrator, a smart contract (Figure \ref{fig:j1}). It transmits information on the use case, reward type and amounts, as well as federation specifics such data scheme, epochs, aggregation strategy and incentive method. The Orchestrator then creates a smart contract for the specific Scenario. The Orchestrator acts as a logbook, keeping track of every federation it has created, enabling prospective clients to choose a Scenario they might want to participate in.

\begin{figure}[h]
\includegraphics[width=8cm]{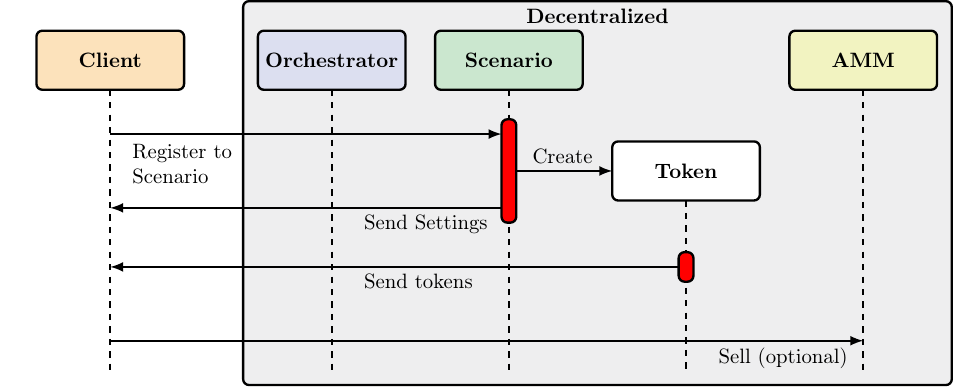}
\caption{A client joins the Scenario to participate in the federation.}
\label{fig:j2}\end{figure}
To join a federation (Figure \ref{fig:j2}), a client contacts the specific Scenario to set up involvement in the federation. During this exchange, the client can choose to create a fixed amount of tokens in their name. The token is tied to the specific client and Scenario, a federation with $n$ clients will contain $n$ tokens at most. Holders of tokens will later receive shares of the specific client's reward, proportional to the amount of tokens they own, just as a shareholder of a company would receive dividends proportional to their held shares. The client receives specific instructions for the setup of the federation, as well as all of their created tokens. They can choose to put their tokens up for sale on the automated market place, or to keep the tokens and collect the full reward.

\begin{figure}[h]
\includegraphics[width=8cm]{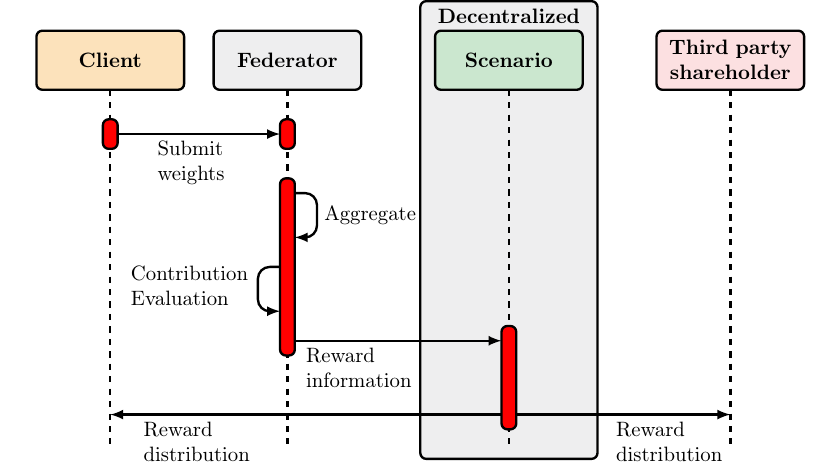}
\caption{The Federated Learning process and subsequent reward payout.}
\label{fig:j3} \end{figure}

Each round, the Federator receives model updates from clients (Figure \ref{fig:j3}), aggregates them and sends back the new global model. Then, the Federator calculates each client's contribution and sends them, along with the round's reward, to the Scenario. Holding ownership information of all tokens, the Scenario splits up the rewards and sends each token holder their respective share of the total reward.
This process repeats at each round: The server determines rewards for all clients and the Scenario sends proportional payments to shareholders.

The Orchestrator and the AMM are not tied to a single federation. On the contrary, tokens from different Scenarios with differing use cases, client types and incentive schemes may be listed on an AMM. Thus, a single company participating in several federations can have multiple types of shares listed on a single exchange.

This framework is independent of the incentive mechanism employed, since in all incentive schemes contain a transfer of value from the server to clients. In this instance, the monetary value is instead transmitted to shareholders. This simple change in dynamics does however clear up several problems of a closed incentive scheme, as we will show.

\subsection{Framework Advantages}
In section \ref{sec:restrictions}, we showcased complications which arise when employing a basic reward payout scheme in a federation, namely the closed flow of capital and potentially high upfront costs. Both of these issues are mitigated by our token-based framework.

Our incentive framework opens up the flow of capital as can be seen in the comparison between Figures \ref{fig:curr_inc} and \ref{fig:sha_exc}. Third-party investing is made possible by the sale of tokens on the AMM and subsequent distribution of rewards to token holders by the Scenario. Thus, investors can easily profit off the federation without having to directly finance clients. However, clients collect the proceeds of the sales of their tokens, enabling them to raise capital off of expected future rewards. Reinvesting this capital with the goal of increasing their contribution to the federation increases token-holder value.

Additionally, our framework allows the sale of tokens at any moment. Joiners to a Scenario have the possibility of selling their created tokens at the start, raising capital to purchase computational equipment and set up participation. We can thus mitigate the impact of upfront cost a federation has on prospective clients. As opposed to a company going public, tokens do not come with voting rights and control over the company, clients only waive their right to future rewards by selling tokens.

The possibility of directly investing in clients introduces a few additional benefits. An investor which sees value in a specific use case can directly profit off the federation by buying tokens in all involved clients. If an investor is confident in the quality of the data held by a specific client, they have the possibility of buying their tokens across all Scenarios the client is a part of, increasing their exposure to the client.
Similarly, clients have the option of buying other participants' tokens. Thus, if a client is convinced by a use case, they can expose their own exposure by buying competitors' tokens, and collecting a bigger portion of the rewards.

Additionally, investors can choose their risk levels: Directly investing in a client which participates in high-risk use cases, an investor runs the risk of their capital being tied to those use cases. By only buying the client's tokens in a specific federation, the investor shields their investments from these federations, while still being able to fund the client.

Since participants have the option of partaking in the learning of process of several Scenarios, they can boost their overall reputation by showing reliable participation  and consistent contribution earnings. This can be an excellent tool to assess a participants' ability to provide valuable data, leading to higher investment interest and increased token prices.

\subsection{Pricing}
Determining asset prices is an important task, as it determines the profitability of investments. As supply and demand drive token prices in an automated market maker, we define the trading function such that each asset is, initially, distributed equally. Hence, each client creates an equal amount of tokens when joining the Scenario. This allows a free market to determine fair token values by themselves via third party investors, driving up prices through demand. 
In literature, different types of reward distributions have been studied: A federator has the possibility of establishing a fixed reward upfront, similarly to a bounty \cite{pandl2022reward}, but rewards may also be changing in value and paid out continuously\cite{yang2017designing}. 

If a \textbf{pre-established reward} amount $R$ is set by the Federator, clients' expected per-round reward can be determined mathematically. With $T$ the total amount of rounds until the reward is fully paid out, and $n$ is the amount of clients in the federation, a client's expected per-round reward $E_R^t$ is $\frac{R}{T \times n}$, with a total expected per-client reward of $\frac{R}{n}$. In such a scenario, each round, tokens lose value proportionally to $T$, since the total reward left diminishes each round, and they become worthless after the last round.

On the other hand, rewards may change between rounds, presumably due to profit generated by Federator through selling the model. In this case, expected reward calculations include an additional uncertainty parameter, namely the expected profit generated. Additionally, federation which do not include a predetermined total reward can be open-ended, meaning that there is no maximum number of rounds. An expected total per-client reward can theoretically be defined as $E_R = \frac{1}{n}\sum_{t=0}{R_t} dt$, where $R_t$ is the total reward at round t. However, as per-round rewards are an unknown quantity, other methods, such as federation-adapted price-to-earnings ratios may be better suited to asses token prices.

Of course, a client's reward may be substantially bigger than the expected per-round reward, as actual reward payouts depend additionally on contribution assessments. As showcased in Figure \ref{img:icaif-prices}, rewards fluctuate considerably between rounds when employing a non-linear and non-equal pricing method. This gives investors the possibility of finding undervalued tokens, enabling speculation. Events such as a client announcing the acquisition of new data sources, a client's unexpected substantial contributions, or the dropping out of a similar competitor, have the potential as serving as catalysts for a token's value.

\section{Conclusion and Future work}\label{sec:con}

This work proposes a novel incentive distribution method using client-specific tokens as a means of investment in federated learning. Our framework addresses the limitations of current incentive mechanisms by introducing a more open financial system, allowing third-party investment and mitigating upfront costs for participants in the learning process by providing them with a novel data monetisation approach. Our approach provides a more inclusive environment, potentially attracting new actors and a wider range of data sources, ultimately enhancing the effectiveness of the overall federated learning process. Additionally, enabling speculation on the value of client tokens could provide valuable insights into the participants' internal data quality.

Thanks to the integration of the proposed framework to a decentralized finance platform, it offers a flexible and adaptive reward system that can dynamically respond to the evolving needs and contributions of participants. Participants can choose to raise more capital through the sale of their tokens or to monetise their data through future contribution earnings. This flexibility is crucial for maintaining participant engagement and incentivizing continuity in their participation.

Future work will involve a more detailed analysis of the economic viability of the proposed framework, as well as in-depth exploration of potential regulatory considerations surrounding the issuance and trading of these client-specific tokens. Additionally, implementing the framework in a specific blockchain environment such as Ethereum and deploying a proof of concept to explore the effect of different pricing strategies would provide further insights into our framework. Finally, further research into the potential integration of advanced cryptographic techniques to enhance security and trust within the framework could further solidify the framework's applicability in privacy-sensitive sectors.

\begin{acks}
We would like to thank the anonymous reviewers for their helpful comments. This paper was funded in part by the Luxembourg National Research Fund (FNR) under grant number 18047633.
\end{acks}

\bibliographystyle{ACM-Reference-Format}
\bibliography{main}

\appendix

\end{document}